\documentclass[10pt,twocolumn,letterpaper]{article}

\usepackage[pagenumbers]{cvpr} 
\usepackage[dvipsnames]{xcolor}
\definecolor{cvprblue}{rgb}{0.21,0.49,0.74}
\usepackage[pagebackref,breaklinks,colorlinks,citecolor=cvprblue]{hyperref}

\usepackage{multirow}
\usepackage{wrapfig}
\usepackage{subcaption}
\usepackage{soul}
\usepackage[ruled]{algorithm2e} 

\newtheorem{theorem}{Theorem}
\newcommand\blfootnote[1]{%
  \begingroup
  \renewcommand\thefootnote{}\footnote{#1}%
  \addtocounter{footnote}{-1}%
  \endgroup
}

\title{Off-the-shelf ChatGPT is a Good Few-shot Human Motion Predictor}

\author{Haoxuan Qu$^{1\dag}$
~~~ Zhaoyang He$^{1\dag}$
~~~ Zeyu Hu$^{2}$
~~~ Yujun Cai$^{3}$
~~~ Jun Liu$^{1\ddag}$ \\
\textsuperscript{1}Singapore University of Technology and Design ~~ \\
\textsuperscript{2}LightSpeed Studios, Tencent ~~ \\
\textsuperscript{3}Nanyang Technological University \\
{\tt\small \{haoxuan\_qu,zhaoyang\_he\}@mymail.sutd.edu.sg, zeyuhu@global.tencent.com}\\ 
{\tt\small  yujun001@e.ntu.edu.sg, jun\_liu@sutd.edu.sg } \\
}

\begin{document}
\maketitle

\blfootnote{\dag~Equal contribution;~~\ddag~Corresponding author} 

\begin{abstract}
To facilitate the application of motion prediction in practice, recently, the few-shot motion prediction task has attracted increasing research attention. 
Yet, in existing few-shot motion prediction works, a specific model that is dedicatedly trained over human motions is generally required. 
In this work, rather than tackling this task through training a specific human motion prediction model, we instead propose a novel \textbf{FMP-OC} framework.
In FMP-OC, \textit{in a totally training-free manner}, we enable \textbf{F}ew-shot \textbf{M}otion \textbf{P}rediction, which is a non-language task, to be performed directly via utilizing the \textbf{O}ff-the-shelf language model \textbf{C}hatGPT.
Specifically, to lead ChatGPT as a language model to become an accurate motion predictor, in FMP-OC, we first introduce several novel designs to facilitate extracting implicit knowledge from ChatGPT. Moreover, we also incorporate our framework with a motion-in-context learning mechanism. Extensive experiments demonstrate the efficacy of our proposed framework.
\end{abstract}

\section{Introduction}

Human motion prediction aims to forecast future human motion sequences based on past observed ones. It is relevant to a variety of applications, such as human-robot interaction \cite{koppula2015anticipating}, human tracking \cite{gong2011multi}, and autonomous driving \cite{levinson2011towards}. 
In the past few years, few-shot human motion prediction \cite{gui2018few,zang2021few,zang2022few,drumond2023few} has received increasing research attention, with the notice that requiring only a few motion samples per each newly appearing human action can largely facilitate the application of human motion prediction in the real world. 
To perform few-shot motion prediction, existing methods \cite{gui2018few,zang2021few,zang2022few,drumond2023few} typically construct a two-step pipeline: 
(1) On a large set of motion samples belonging to certain base human actions, a specific model is first trained through a dedicatedly-designed base training process for the motion prediction task. (2) The specific model obtained in step (1) is then adapted to the few motion samples belonging to each newly appearing human action. Considering that in this pipeline, it is necessary to train a specific motion prediction model, such a pipeline may sometimes be a bit inconvenient in certain real-world applications.

On the other hand, we notice that recently, pre-trained in a general manner, the off-the-shelf ChatGPT model \cite{ChatGPT} has been found to contain rich implicit knowledge \cite{qu2023lmc}. 
Holding such rich knowledge, via provided with a few demonstrations of a certain task as the context in the prompt (i.e., in-context learning \cite{brown2020language}), ChatGPT has already been successfully utilized to handle various language-related tasks in a few-shot manner, such as translation \cite{zhu2023multilingual} and question answering \cite{guan2024mfort}, conveniently without any training or tuning.
Motivated by this, we here are wondering, rather than training a specific motion prediction model, from a novel perspective, \textit{if we can directly perform motion prediction in an effective yet totally training-free manner, via utilizing the off-the-shelf language model ChatGPT to be the human motion predictor?}

In general, ChatGPT possesses certain qualities that are useful for human motion prediction: (1) ChatGPT is pre-trained over an extremely large corpus that generally includes 
rich descriptions w.r.t. how human beings behave and move when performing different actions (i.e., rich human motion descriptions over different actions) \cite{brown2020language}. Thus, ChatGPT can hold rich knowledge of human motions. (2) Besides sentences in natural languages, the training corpus of ChatGPT generally also contains rich programming data \cite{feng2023layoutgpt}. Pre-trained over such rich programming data that typically involves both numbers and coordinates, ChatGPT can also be equipped with certain numerical understanding abilities.
Here, we notice that in the human motion prediction task, each human pose in a motion sequence can essentially be interpreted as a special combination of joint coordinates that involve human motion prior. Intuitively, this makes it possible for harnessing the off-the-shelf ChatGPT model possessing the above two qualities to perform human motion prediction in a training-free manner, via passing the past observed motion sequence as a list of joint coordinates to ChatGPT.

\begin{figure}[t]
\centering
\includegraphics[width=\columnwidth]{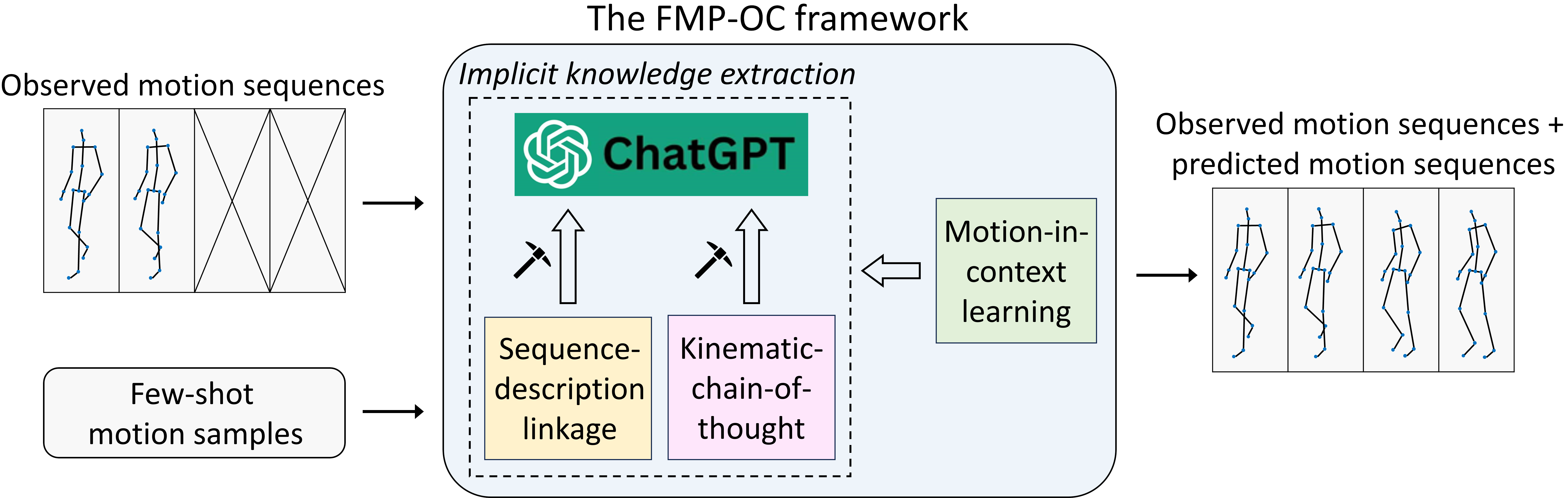}
\vspace{-0.55cm}
\caption{Illustration of our FMP-OC framework. As shown, we first incorporate our framework with two designs (in yellow and pink respectively) to extract (mine) implicit knowledge from ChatGPT effectively. Moreover, we also incorporate our framework with a motion-in-context learning mechanism (in green), which can further increase ChatGPT's familiarity with the human motion prediction task.}
   \label{fig:intro}
    \vspace{-0.5cm}
\end{figure}

However, despite the success of ChatGPT in handling various language-related tasks in a few-shot manner, enabling ChatGPT to perform accurate few-shot human motion prediction without any further training can still be non-trivial due to the following challenges: (1) Despite being pre-trained over both natural language sentences and programming data, ChatGPT naturally has not been specifically pre-trained over human motion sequences. In other words, the human motion sequence is not in a format that is ``familiar'' to ChatGPT. (2) The knowledge contained in ChatGPT is ``implicit'' \cite{wang2023does,ashby2023personalized}. This implies that, while ChatGPT can hold certain knowledge that can be helpful to the human motion prediction task, how to successfully extract such implicit knowledge from ChatGPT can remain challenging. 
To tackle the above challenges, in this work, we propose a novel framework named \textbf{F}ew-shot \textbf{M}otion \textbf{P}rediction directly via \textbf{O}ff-the-shelf \textbf{C}hatGPT (\textbf{FMP-OC}), which represents the first method that can perform few-shot human motion prediction in a totally training-free manner, via directly leveraging the off-the-shelf ChatGPT model to be the motion predictor. We illustrate our framework in Fig.~\ref{fig:intro}, and outline our framework as follows.

Overall, to enable ChatGPT as a language model to become an accurate motion predictor, in our framework, we need to ``familiarize'' ChatGPT with human motion sequences that it is originally ``unfamiliar'' with. 
To achieve such familiarization, motivated by that ChatGPT can hold rich implicit knowledge over human motions, in FMP-OC, we first propose two designs to facilitate extracting useful implicit knowledge from ChatGPT. Specifically, these two designs include (1) guiding ChatGPT to link motion sequences with motion descriptions that it is more ``familiar'' with (i.e., performing sequence-description linkage), and (2) introducing ChatGPT with a kinematic-chain-of-thought design.

Furthermore, we observe that, in few-shot motion prediction, over any newly appearing human action, we can have a few shots of supporting motion samples from this new action, as well as a large set of motion samples from certain base actions, to be available.
Motivated by this, for ChatGPT to become an accurate motion predictor, in our framework, besides urging ChatGPT to use its own implicit knowledge, we also introduce a motion-in-context learning mechanism for guiding ChatGPT to seek help from the above-mentioned available motion samples. Specifically, in this mechanism, through a motion sample pre-selection stage, we first pre-select a subset of motion samples from all available ones that can benefit ChatGPT to a large extent. We then demonstrate these motion samples to ChatGPT effectively in a practice-exam-based demonstration manner. Through the above process, FMP-OC can finally output the forecasted motion sequences directly from ChatGPT's answers reliably.

The contributions of our work are as follows.
(1) We propose a novel few-shot human motion prediction framework named FMP-OC. To the best of our knowledge, this work is the first exploration on performing few-shot human motion prediction, which is a non-language task, via directly utilizing the off-the-shelf large language model ChatGPT.
(2) We introduce several designs in our framework to familiarize ChatGPT with motion sequences and facilitate ChatGPT in becoming an accurate motion predictor.
(3) In a totally training-free manner, FMP-OC achieves state-of-the-art performance on the evaluated benchmarks.

\begin{figure*}[t]
\centering
\includegraphics[width=0.85\textwidth]{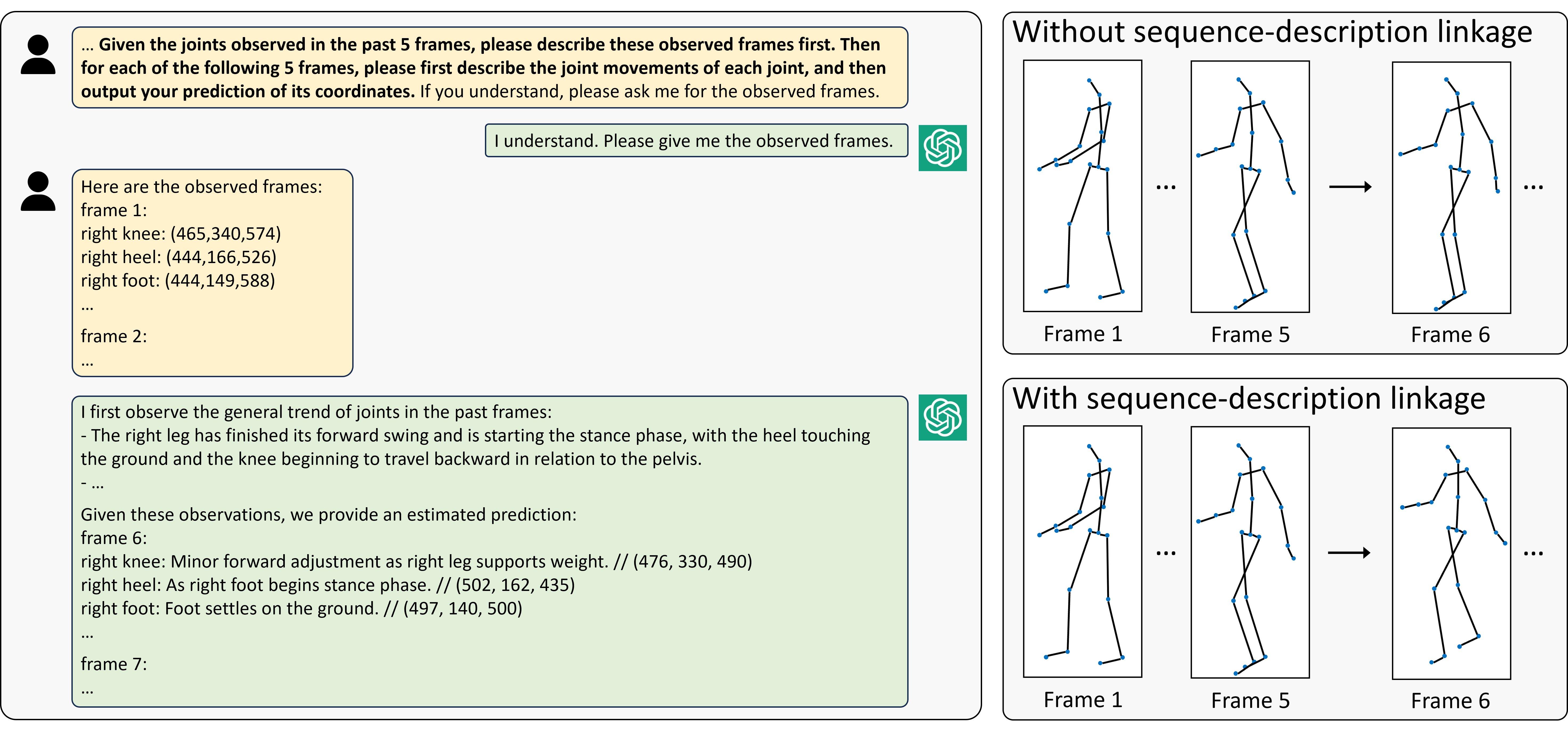}
\vspace{-0.2cm}
\caption{Illustration of the language command for ChatGPT when human poses in the motion sequence (in each frame) are guided to be linked to their corresponding motion descriptions. In the demonstrated example, the length of the past observed motion sequence $L$ is set to 5, and the length of the future motion sequence $J$ is set to 5.}
\label{fig:link}
\vspace{-0.2cm}
\end{figure*}

\section{Related Work}

\textbf{Human Motion Prediction.} Due to the wide range of applications, human motion prediction has received lots of research attention \cite{fragkiadaki2015recurrent,gui2018adversarial,cai2020learning,xu2023auxiliary,gui2018few,zang2021few,zang2022few,drumond2023few,gao2023decompose,guo2023back,aksan2021spatio,li2022skeleton,ma2022progressively,sofianos2021space,dang2021msr,barquero2023belfusion,mao2021generating,martinez2021pose,tang2024temporal}. In the early days, most works in this area focused on the setting where sufficient training motion samples are given for each type of human action. Among them, Fragkiadaki et al. \cite{fragkiadaki2015recurrent} proposed the Encoder-Recurrent-Decoder model which incorporates the LSTM structure with an encoder-decoder network. After that, Cai et al. \cite{cai2020learning} proposed to perform human motion prediction using a transformer architecture, and considered both the spatial correlations and temporal smoothness of human motions at the same time. Later on, to lead the forecasted human motions to be more realistic, 
Mao et al. \cite{mao2021generating} proposed to further seek help from a pre-trained normalizing flow model.
More recently, Xu et al. \cite{xu2023auxiliary} proposed to facilitate the prediction of human motions via jointly learning this task with two auxiliary tasks including denoising and masking prediction. Besides the above works that generally focus on tackling only the motion prediction task, recently, several methods \cite{jiang2023motiongpt,zhang2023motiongpt} have also been proposed to handle various different types of motion-related tasks in a unified manner, via training or finetuning a large-scale motion model on a huge amount of motion samples. 

Besides the above setting, few-shot human motion prediction \cite{gui2018few,zang2021few,zang2022few,drumond2023few} has also gained much research attention. Specifically, Gui et al. \cite{gui2018adversarial} proposed Proactive and Adaptive Meta-Learning (PAML) that combines model-agnostic meta-learning \cite{finn2017model} with regression networks, and made the first attempt at performing few-shot human motion prediction. Later on, MoPredNet, which creates a new set of parameters for each new category of human motions, was proposed by Zang et al. \cite{zang2021few}. After that, Drumond et al. \cite{drumond2023few} further proposed GraphHetNet and leveraged the underlying graph structure of human skeletons.

Different from these approaches, in this work, from a novel perspective, we investigate how to perform few-shot human motion prediction in a totally training-free manner, utilizing the off-the-shelf ChatGPT model.
To achieve this, we introduce several novel designs in our proposed FMP-OC framework to familiarize ChatGPT with human motions. By doing so, we can directly acquire human motion predictions from ChatGPT's output.

\noindent\textbf{Large Language Models.} Recently, various different large language models, such as ChatGPT \cite{ChatGPT}, have been proposed. Pre-trained over a tremendously large training corpus that generally contains both natural sentences and programming data, these large language models have been found to contain rich implicit knowledge \cite{qu2023lmc}, and have thus been applied in various different tasks \cite{zhu2023multilingual,guan2024mfort,firdous2023openai,shi2023language,shao2023prompting,lian2023llm}, such as code interpretation \cite{firdous2023openai} and event prediction \cite{shi2023language}. In this work, we design a novel framework, in which rather than optimizing a new motion prediction model, for the first time, we enable few-shot human motion prediction, which is a non-language task, to be performed via the off-the-shelf language model ChatGPT.

\section{Proposed Method}

With respect to the few-shot motion prediction task, for each newly appearing human action $a$, $N$ motion samples $\{(m^{n,a}_{obs}, m^{n,a}_{fut})\}^{N}_{n=1}$ from this action category are first provided as the supporting set, where $m^{n,a}_{obs}$ represents the past observed motion sequence of the $n$-th sample, and $m^{n,a}_{fut}$ represents the ground-truth future motion sequence of the $n$-th sample. 
Then based on the provided supporting set from $a$, given a newly observed motion sequence $m^{test,a}_{obs}$ from the same action category, the goal of this task is to forecast its corresponding future motion sequence $m^{test,a}_{fut}$. 
To perform this task conveniently, in this work, inspired by that the off-the-shelf language model ChatGPT can hold rich implicit knowledge over human motions, we aim to directly utilize ChatGPT to be the human motion predictor, in a totally training-free manner, through a novel framework \textbf{FMP-OC}. 
As shown in Fig.~\ref{fig:intro}, in our framework, to acquaint ChatGPT as a language model with human motion sequences, we first introduce two designs to facilitate extracting implicit knowledge from ChatGPT. Moreover, a new type of in-context learning mechanism (i.e., motion-in-context learning) is also incorporated into our framework for further increasing ChatGPT's familiarity with the human motion prediction task.

Below, we first describe the two designs for more effectively extracting implicit knowledge from ChatGPT. We then discuss the motion-in-context learning mechanism.

\subsection{Implicit Knowledge Extraction}
\label{sec:IKE}

In our framework, we aim to perform motion prediction directly via utilizing ChatGPT. However, since motion sequences are not in a format that ChatGPT as a language model is ``familiar'' with, naively passing past observed motion sequences to ChatGPT and requiring it to forecast future motion sequences can lead to its understanding difficulty, and thus result in poor motion prediction results. This makes familiarizing ChatGPT with motion sequences an important step in our framework. To achieve this, with the observation that ChatGPT already implicitly holds certain knowledge that can be helpful in understanding human motions, one potential way can be to extract such implicit knowledge from ChatGPT. Specifically, we introduce our framework with the following designs to facilitate extracting such implicit knowledge from ChatGPT. 
These designs include 
(\romannumeral1) guiding ChatGPT to link motion sequences with motion descriptions that it is more ``familiar'' with (i.e., performing sequence-description linkage), 
and (\romannumeral2) introducing ChatGPT with a kinematic-chain-of-thought design.

\subsubsection{Sequence-description linkage design.}

As for the sequence-description linkage design in (\romannumeral1), we get inspiration from that, despite the unfamiliarity with human motion sequences, ChatGPT has seen a large number of human motion descriptions (i.e., descriptions w.r.t. how human beings behave and move when performing different actions) during its pre-training stage. 
Considering this, rather than directly asking ChatGPT to predict future motion sequences, we aim to instruct ChatGPT to first link motion sequences with motion descriptions.
By doing so, we can enable ChatGPT to better leverage its rich knowledge over motion descriptions, to tackle the motion sequence prediction task that it is relatively ``unfamiliar'' with. 

Specifically, we find that one effective way to guide ChatGPT to build the linkage is to instruct ChatGPT through the following command: ``Given the joints observed in the past [$L$] frames, please describe these observed frames first. Then for each of the following [$J$] frames, please first describe the joint movements of each joint, and then output your prediction of its coordinates.'', where $L$ represents the length of the past observed motion sequence, and $J$ represents the length of the future motion sequence that is required to be forecasted. 
In Fig.~\ref{fig:link}, we illustrate an example when the above command is passed to ChatGPT. As shown, by formulating the command to ChatGPT in this way, ChatGPT is first demanded to link the past observed motion sequence with its corresponding motion descriptions. Then during predicting the human poses in the future motion sequence, ChatGPT can utilize its motion description predictions that it has more experts at to facilitate its prediction of the motion sequences (joint coordinates).
In particular, as shown on the right side of Fig.~\ref{fig:link}, when not guided with the sequence-description linkage, ChatGPT predicts the human pose in frame 6 in low quality by almost copying the last human pose (in frame 5), while with the help of motion descriptions of each joint, ChatGPT predicts the next human pose much more reasonably. This demonstrates the effectiveness of linking motion sequences with motion descriptions.

\subsubsection{Kinematic-chain-of-thought design.} 

We also internally leverage the structural connectivity of the human skeleton in the motion sequence, via a kinematic-chain-of-thought design. Specifically, we get inspiration from \cite{cai2020learning} that, 
for human beings who typically hold a fixed spatial skeleton structure, based on the kinematic chains of human skeletons, their motions naturally would propagate in a fixed order from the center part of the body trunk to the peripheral limbs. In other words, when a person performs a certain motion, the movements of the body joints in the peripheral limbs of the human skeleton naturally are dependent on the movements of their parent body joints in the more central part of the human skeleton. 
For example, across two consecutive frames, if a person tries to move his \textit{right shoulder}, such a movement would then sequentially affect the location of his \textit{right elbow}, and the location of his \textit{right wrist}.

Inspired by the above that human motions naturally propagate through different body joints in a fixed order, in our framework, we aim to encourage ChatGPT to predict joint coordinates in every human pose in the future motion sequence also in such a central-to-peripheral order following the kinematic chains of the body skeleton. 
To achieve this, for every human pose of the past observed motion sequence, rather than demonstrating its body joint coordinates to ChatGPT in a random order (like in Fig.~\ref{fig:link}), we propose to demonstrate body joint coordinates to ChatGPT in a central-to-peripheral order instead. 
By doing so, ChatGPT can then be hinted to predict the body joint coordinates of each future human pose also in such an order, simulating the natural motion propagation order of the human skeleton. Making joint coordinate predictions in such a way, ChatGPT can then be encouraged to leverage the internal structural connectivity of human skeletons to make its predictions over joint coordinates in human poses more naturally and reasonably. 
We also involve a more detailed discussion of both the kinematic chains and the central-to-peripheral order in supplementary.

\subsection{Motion-in-context Learning}
\label{sec:icl}

To guide ChatGPT in becoming an accurate motion predictor, above we focus on facilitating ChatGPT in effectively extracting its own implicit knowledge. Here from another perspective, we notice that, in few-shot motion prediction, over an arbitrary newly appearing human action $a$, certain other resources besides ChatGPT's own knowledge generally are also available. These resources include the $N$ supporting motion samples of $a$, and the large set of motion samples from the base training set $S$. Considering this, in our framework, for enabling ChatGPT to perform motion prediction more accurately, besides urging ChatGPT to better utilize its own implicit knowledge, we also aim to find a way, through which ChatGPT can seek help from those available motion samples effectively. Specifically, to achieve this, a naive way can be to demonstrate all available motion samples as the context in the prompt to ChatGPT (i.e., perform in-context learning leveraging all available motion samples), with the notice that in-context learning with more samples theoretically tends to be more beneficial to the large language model \cite{xie2021explanation}. Unfortunately, as shown in the previous study \cite{long_context}, in practice, super-long contexts can trigger catastrophic forgetting of ChatGPT and result in a significant performance drop.

Considering this, in our framework, we instead propose to guide ChatGPT to seek help from the available motion samples through a novel motion-in-context learning mechanism. 
Specifically, in this mechanism, before performing in-context learning over any action, through a \textbf{pre-selection stage}, we first pre-select $P$ motion samples from the base training set that, when utilized during in-context learning, can largely deepen ChatGPT's understanding over the motion prediction task. 
Moreover, during in-context learning over an arbitrary action $a$, to make good use of the $P$ pre-selected motion samples together with the $N$ supporting samples from $a$, rather than simply passing ChatGPT with the concatenation of these $P + N$ samples following the typical in-context learning in the language-related tasks \cite{brown2020language}, we further propose to demonstrate these samples to ChatGPT \textbf{in a practice-exam-based manner}.
Below, we go through the pre-selection stage and the practice-exam-based sample demonstration process in our motion-in-context learning mechanism.

\subsubsection{Pre-selection stage.}

With the notice that passing ChatGPT with super-long context can lead it to suffer from the catastrophic forgetting problem, rather than demonstrating the whole base training set $S$ to ChatGPT, in our motion-in-context learning mechanism, we instead first incorporate a motion sample pre-selection stage. Specifically, in this stage, assuming that we can only demonstrate $P$ motion samples from the base training set to ChatGPT during in-context learning, where $P$ is a hyperparameter, we aim to select a proper subset of $P$ samples, such that their demonstration during in-context learning can largely deepen ChatGPT's understanding of the motion prediction task.

\ul{Motivation behind the construction of this stage.} 
To construct such a pre-selection stage, we get inspiration from the previous works \cite{wang2023large,devroye2013probabilistic}. In these works, via interpreting in-context learning as a latent topic model of a certain task, it is theoretically analyzed that, under a mild assumption, if the samples used during in-context learning can represent the task precisely, in-context learning can reach its Bayesian optimal state and thus be most beneficial to the large language model.
Motivated by this, here in our pre-selection stage, we aim to select the $P$ motion samples from the base training set in a way, such that they can well-represent the motion prediction task. Specifically, below, as a preparation, we first describe that, given a subset ($S_{sub}$) of $P$ motion samples from the base training set $S$, how we can measure its representativeness over the motion prediction task. We then discuss the motion sample selection algorithm we use during this stage, based on the described representativeness measurement.

\underline{Representativeness measurement.} Here, we aim to measure the representativeness of $S_{sub}$ over the motion prediction task. However, as motion prediction by itself is just an abstract topic (concept), it can be hard to conduct such a measurement directly. Thus, with the notice that the base training set $S$ can be regarded as a large set of motion samples sampled from the abstract motion prediction topic, we here instead approximate the representativeness of $S_{sub}$ by measuring how well this subset can represent $S$. 
Specifically, we first notice that, in previous works \cite{kempe2003maximizing,zhang2023ideal,kumar2021information}, the information diffusion process has been widely used to measure the representativeness of a certain subset over the whole set, e.g., the representativeness of a subset of social units over the entire social graph.

Motivated by these previous works, we here measure the representativeness of $S_{sub}$ over $S$ in an information-diffusion-based manner via the following three steps.
\textbf{(1) Distance measurement.} To reformulate $S$ like a (social) graph, given $m_i = (m^{i,S}_{obs}, m^{i,S}_{fut})$ and $m_j = (m^{j,S}_{obs}, m^{j,S}_{fut})$ the $i$-th motion sample and the $j$-th motion sample in $S$, as a preparation, we first need to measure the distance $d(m_i, m_j)$ between $m_i$ and $m_j$. To better achieve this, we get inspiration from \cite{li2022skeleton} that, in a motion sequence, joints in different body parts of the human pose often can be under different motion patterns. Thus, besides measuring the global L2 distance between $m_i$ and $m_j$, we here also measure five local L2 distances between five different body parts of $m_i$, and the corresponding five body parts from $m_j$. 
These five body parts include left arm, right arm, left leg, right leg, and body truck. 
After measuring the global L2 distance between $m_i$ and $m_j$ and the five local L2 distances between $m_i$ and $m_j$, we finally measure $d(m_i, m_j)$ as the average of the above six distances. \textbf{(2) Matrix formulation.} After measuring the distance between every pair of motion samples in $S$ using the distance function $d(\cdot, \cdot)$ defined in step (1), we then leverage these measured distances to form the similarity matrix $M \in \mathbb{R}^{|S|\times|S|}$, where the $[i,j]$-th entry of $M$ represents the similarity of $m_i$ and $m_j$, calculated as $\frac{1}{d(m_i, m_j)}$. After deriving the matrix $M$, we formulate the matrix $M_N$ from $M$ via normalizing every row of $M$. Note that we only need to construct $M_N$ once, and it can then be repeatedly used for measuring the representativeness of different subsets of $S$. \textbf{(3) Representativeness measurement.} Once we construct the matrix $M_N$, we measure the representativeness $r$ of $S_{sub}$ over $S$ via Alg.~\ref{algorithm_1}. Intuitively, Alg.~\ref{algorithm_1} can be understood as a recursive diffusion process spreading from the motion samples in $S_{sub}$ towards the whole base training set $S$. At the end of such a diffusion process, the representativeness $r$ is then measured as the total number of samples that are spread over during the process. A more detailed explanation of Alg.~\ref{algorithm_1} is also provided in supplementary. Note that besides subsets with size $P$, Alg.~\ref{algorithm_1} can also be used to measure the representativeness of subsets with other sizes.

\begin{algorithm}[t]
\footnotesize
 \nl {\bf Input:} The subset $S_{sub}$, the base training set $S$, and the matrix $M_N$.\\
 \nl {\bf Output:} The representativeness $r$ of $S_{sub}$. \\
 \textcolor{blue}{// Initialize the set of motion samples $S_{spread}$ that are spread over in the diffusion process to be $S_{sub}$.} \\
 \nl Set $S_{spread} = S_{sub}$. \\
 \nl Set $S_{cur} = S_{sub}$, and $S_{next} = \varnothing$. \\
 \textcolor{blue}{// In each of the while loop, we try to spread from $S_{cur}$ to motion samples in $S \setminus S_{spread}$, i.e., motion samples that are not yet spread over.}\\
 \nl \While{$S_{cur} \neq \varnothing$}  {
     \nl \For{every motion sample in $S_{cur}$} {
         \nl Denote the row of the matrix $M_N$ that corresponds to the current sample $r_N$. \\
         \nl \For{every element in the row $r_N$} {
             \nl Denote $rand$ a random number generated between 0 and 1. \\
             \nl Denote current element the $i$-th element of $r_N$. \\
             \textcolor{blue}{// A motion sample in $S \setminus S_{spread}$ is spread over in the current while loop if it is close enough to a motion sample in $S_{cur}$.}\\
             \nl \uIf{the $i$-th element of $S$ is in $S \setminus S_{spread}$ \textbf{AND} $r_N[i] > rand$ } {
                 \nl Add the $i$-th element of $S$ into both $S_{spread}$ and $S_{next}$.
            }
        }
    }
     \nl Set $S_{cur} = S_{next}$, and $S_{next} = \varnothing$. \\
}
 \textcolor{blue}{// We finally calculate $r$ as the number of samples that are spread over throughout the diffusion process (i.e., the size of $S_{spread}$).} \\
 \nl {\bf return} $r = |S_{spread}|$
\scriptsize
\caption{Measuring the representativeness $r$ of $S_{sub}$ through recursive diffusion.}
\label{algorithm_1}
\end{algorithm}
\setlength{\floatsep}{5pt}

\begin{algorithm}[t]
\footnotesize
 \nl {\bf Input:} The expected size $P$ of the subset, the base training set $S$, and the matrix $M_N$.\\
 \nl {\bf Output:} The found subset $S_{sub}$. \\
 \nl Set $S_{sub} = \varnothing$, and denote Alg.~\ref{algorithm_1} as $f_{alg1}(\cdot)$ \\
 \nl \While{$|S_{sub}| < P$} {
     \nl Find the motion sample $m$ from $S \setminus S_{sub}$ that leads to maximum $f_{alg1}(S_{sub} \cup m)$. \\
     \nl Add the found sample $m$ into $S_{sub}$.
}
 \nl {\bf return} $S_{sub}$
\scriptsize
\caption{The greedy algorithm to practically find $S_{sub}$ in our framework.}
\label{algorithm_2}
\end{algorithm}
\setlength{\textfloatsep}{5pt}

\underline{How we find $S_{sub}$ practically.} Once we formulate how to measure the representativeness $r$ given a subset $S_{sub}$, to find the subset of size $P$ that can well-represent the motion prediction task, a naive way can be to exhaust all subsets of $S$ that contains $P$ motion samples, and find the subset with the highest $r$ among them. Yet, during our experiment, we find that for the base training set $S$ that typically contains a large number of motion samples, such an exhaustion as an NP-hard problem can generally be computationally impractical. Considering this, we here instead find $S_{sub}$ practically using the greedy algorithm demonstrated in Alg.~\ref{algorithm_2}. In Alg.~\ref{algorithm_2}, $S_{sub}$ is initialized to be an empty set. Then until the size of $S_{sub}$ reaches $P$, during every step, the motion sample from $S \setminus S_{sub}$ that, when added to $S_{sub}$, leads to the highest representativeness $r$, is added to $S_{sub}$. Below, we provide a theoretical analysis over the effectiveness of Alg.~\ref{algorithm_2}.

\begin{theorem}
\label{thm:theorem_2}
Define the optimal subset $S^*_{sub}$ the subset of the base training set $S$ that has the highest representativeness over all subsets of $S$ that contains $P$ motion samples. Denote $r^*$ the representativeness of $S^*_{sub}$. The relationship between the representativeness $r$ of the subset $S_{sub}$ found in Alg.~\ref{algorithm_2} and $r^*$ of $S^*_{sub}$ can be written as
\begin{equation}\label{eq:theorem_2}
\setlength{\abovedisplayskip}{3pt}
\setlength{\belowdisplayskip}{3pt}
\begin{aligned}
r \geq \big( 1 - (1 - \frac{1}{P})^P\big) r^*
\end{aligned}
\end{equation}
\end{theorem}

The proof of Theorem~\ref{thm:theorem_2} is provided in supplementary. 
Note that for $1 - (1 - \frac{1}{P})^P$ on the right hand side of Eq.~\ref{eq:theorem_2}, it is a monotone decreasing function when $P \geq 1$, and it finally converges to $1 - \frac{1}{e} \approx 63\%$ when $P$ approaches positive infinity. This indicates that, the representativeness $r$ of the subset $S_{sub}$ that can be practically formed via Alg.~\ref{algorithm_2} is guaranteed to be at least lower bounded by a certain relatively large proportion of the representativeness $r^*$ of the optimal subset $S^*_{sub}$. This leads the $P$ motion samples in $S_{sub}$ to keep at least a relatively good representation of the motion prediction task. Thus, our motion-in-context learning mechanism, when equipped with these samples from $S_{sub}$, can be expected to largely deepen ChatGPT's understanding of the motion prediction task. In contrast, as mentioned above, finding $S^*_{sub}$ directly via exhaustion typically requires an impractical number of computations.

\subsubsection{Practice-exam-based sample demonstration.} 

Above we select the $P$ motion samples from the base training set that when utilized during in-context learning, can assist ChatGPT in getting more familiar with the motion prediction task to a large extent. Here, during in-context learning over the newly appearing action $a$, to make good use of these $P$ samples together with the $N$ supporting motion samples from action $a$, we propose to demonstrate these $P+N$ motion samples to ChatGPT during in-context learning in a practice-exam-based manner. As suggested by its name, this proposed practice-exam-based demonstration process is different from how in-context learning \cite{brown2020language} typically demonstrates in-context samples to the large language model, in which both the questions and their corresponding correct answers in the in-context examples are directly passed to the large language model. Instead, in our proposed practice-exam-based process, ChatGPT is required to practice the questions first before being given their corresponding correct answers. This is inspired by the observation that, when students revise for their exams through a practice exam paper, simply reading through the questions and answers of the paper often fails to result in a high-quality revision. Instead, a better revision method for students can be to first practice the questions in the paper, and then compare their own answers with the correct answers to see where they make mistakes and which knowledge points still need to be better reviewed.

Based on this finding, in our framework, we formulate instructions passed through the practice-exam-based demonstration process to ChatGPT as follows: 
(1) ``Below, please first ask me for examples of good prediction.'';
(2) ``Here are the observed frames of an example: [observed]. Now please give your prediction for all joint coordinates of the following frames. Respond `Finished' once you finish.'', where [observed] represents the past observed motion sequence of the $i$-th sample among the $P+N$ motion samples where $i \in \{1, ..., P+N\}$; 
(3) ``Here are the correct answers: [future]. Please check if your predictions and the correct answers are similar. Besides, please also check if your predictions keep limbs the same length across frames (same as the correct answers).'', where [future] represents the corresponding ground-truth future motion sequence of the $i$-th motion sample. 
Note that instructions in (2) and (3) are repeatedly passed to ChatGPT until it has practiced over all the $P+N$ motion samples. By demonstrating these $P+N$ motion samples to ChatGPT like this, instead of directly passing ChatGPT the correct answers of the $P+N$ samples (i.e., the ground-truth future motion sequences), we encourage ChatGPT to practice over its originally ``unfamiliar'' motion prediction task first. Equipped with such a practicing step, ChatGPT can then be encouraged to be more prepared for the motion prediction task.

\subsection{Overall Inference Process}

Above we discuss different parts of our framework. Here, we summarize our framework's overall inference process. Specifically, before entering the inference process, in our framework, we first prepare the subset $S_{sub}$ following the pre-selection stage in Sec.~\ref{sec:icl}. After preparing $S_{sub}$, the inference process of our framework over any action is performed via demonstrating ChatGPT with motion samples in a practice-exam-based manner, and incorporating ChatGPT with designs introduced in Sec.~\ref{sec:IKE} to effectively extract its implicit knowledge. Finally. we retrieve the forecasted motion sequence from ChatGPT's outputs. Note that the whole inference process can be automatically performed with a script.

\begin{table*}[t]
\caption{Results on the Human3.6M dataset. We report MPJPE in millimeters, and lower MPJPE indicates better model performance.}
\vspace{-0.4cm}
\label{Tab:human3.6m}
\resizebox{\textwidth}{!}
{
\begin{tabular}{c|cccccccccc|c|cccccccccc|c} \hline
\multirow{2}{*}{\textbf{Method}} 
& \multicolumn{11}{c|}{Walking} 
& \multicolumn{11}{c}{Eating} \\ 
\cline{2-23}
& 80ms & 160ms & 240ms & 320ms & 400ms & 560ms & 640ms & 720ms & 880ms & 1000ms & Avg. 
& 80ms & 160ms & 240ms & 320ms & 400ms & 560ms & 640ms & 720ms & 880ms & 1000ms & Avg. 
\\ \hline\hline
PAML 
& 36.1 & 53.5 & 65.3 & 72.0 & 79.5 
& 84.6 & 87.6 & 91.6 & 96.1 & 103.0 
& 76.9
& 26.8 & 38.7 & 48.1 & 56.7 & 68.5 
& 81.2 & 90.7 & 93.2 & 97.0 & 99.4 
& 70.0\\
TimeHetNet 
& 29.3 & 35.3 & 49.1 & 58.0 & 64.1
& 75.7 & 81.0 & 87.8 & 90.5 & 92.3 
& 66.3
& 22.8 & 25.6 & 41.4 & 55.8 & 67.3 
& 80.5 & 87.4 & 89.2 & 92.0 & 96.7 
& 65.9 \\
MoPredNet 
& 24.0 & 31.0 & 42.9 & 47.7 & 55.2  
& 63.7 & 67.6 & 73.8 & 76.6 & 78.2 
& 56.1
& 20.0 & 27.2 & 39.9 & 44.2 & 54.0  
& 69.2 & 72.5 & 79.1 & 83.2 & 88.3 
& 57.8\\
GraphHetNet 
& 23.2 & 33.6 & 45.8 & 52.5 & 58.7 
& 65.7 & 67.5 & 69.1 & 71.9 & 72.5 
& 56.1
& 18.6 & 26.6 & 38.8 & 48.5 & 59.2 
& 69.8 & 75.7 & 80.6 & 88.9 & 93.6 
& 60.0\\
\hline
Ours  
& \textbf{16.2} & \textbf{24.9} & \textbf{32.3} & \textbf{38.0} & \textbf{41.8} 
& \textbf{43.9} & \textbf{46.4} & \textbf{47.6} & \textbf{52.7} & \textbf{53.2} & \textbf{39.7}
& \textbf{13.1} & \textbf{20.4} & \textbf{29.6} & \textbf{38.9} & \textbf{46.9} 
& \textbf{57.4} & \textbf{61.9} & \textbf{66.3} & \textbf{72.5} & \textbf{74.7} & \textbf{48.2}\\
\hline
\end{tabular}}
\resizebox{\textwidth}{!}
{
\begin{tabular}{c|cccccccccc|c|cccccccccc|c} \hline
\multirow{2}{*}{\textbf{Method}} 
& \multicolumn{11}{c|}{Smoking} 
& \multicolumn{11}{c}{Discussion} \\ 
\cline{2-23}
& 80ms & 160ms & 240ms & 320ms & 400ms & 560ms & 640ms & 720ms & 880ms & 1000ms & Avg. 
& 80ms & 160ms & 240ms & 320ms & 400ms & 560ms & 640ms & 720ms & 880ms & 1000ms & Avg. 
\\ \hline\hline
PAML 
& 31.1 & 34.2 & 45.5 & 52.7 & 62.8 
& 73.5 & 79.5 & 89.5 & 96.0 & 103.8 
& 66.8
& 36.5 & 45.1 & 53.9 & 70.7 & 84.2
& 100.1 & 107.4 & 121.8 & 127.0 & 128.4 
& 87.5 \\
TimeHetNet 
& 39.0 & 42.9 & 47.3 & 51.9 & 65.3 
& 79.7 & 82.2 & 90.7 & 99.9 & 104.0 
& 70.3
& 37.0 & 48.7 & 62.7 & 77.9 & 89.5
& 113.2 & 122.0 & 129.4 & 135.6 & 139.6
& 95.6 \\
MoPredNet 
& 24.8 & 29.7 & 37.1 & 45.3 & 57.8  
& 67.5 & 71.9 & 80.3 & 90.3 & 98.3  
& 60.3
& 29.8 & 43.3 & 51.5 & 65.7 & 79.0
& 99.3 & 105.5 & 114.5 & 119.3 & 119.6  
& 82.8 \\
GraphHetNet 
& 15.9 & 20.7 & 31.7 & 40.1 & 50.8 
& 62.6 & 65.0 & 71.1 & 81.3 & 95.3 
& 53.5
& 18.2 & 31.6 & 45.6 & 59.2 & 72.5 
& 101.1 & 111.5 & 121.6 & 126.7 & 128.7 
& 81.7\\
\hline
Ours 
& \textbf{10.1} & \textbf{17.3} & \textbf{21.8} & \textbf{26.2} & \textbf{29.7} 
& \textbf{41.6} & \textbf{42.9} & \textbf{47.4} & \textbf{58.3} & \textbf{67.2} & \textbf{36.3}
& \textbf{16.0} & \textbf{31.4} & \textbf{44.0} & \textbf{52.8} & \textbf{63.5} & \textbf{91.4} & \textbf{100.5} & \textbf{109.9} & \textbf{117.7} & \textbf{118.0} & \textbf{74.5}\\
\hline
\end{tabular}}
\vspace{-0.1cm}
\end{table*}

\begin{table*}[t]
\caption{Results on the CMU Mocap dataset. We report MPJPE in millimeters.}
\vspace{-0.4cm}
\label{Tab:cmu}
\resizebox{\textwidth}{!}
{
\begin{tabular}{c|cccccccccc|c|cccccccccc|c} \hline
\multirow{2}{*}{\textbf{Method}} 
& \multicolumn{11}{c|}{Running} 
& \multicolumn{11}{c}{Soccer} \\ 
\cline{2-23}
& 80ms & 160ms & 240ms & 320ms & 400ms & 560ms & 640ms & 720ms & 880ms & 1000ms & Avg. 
& 80ms & 160ms & 240ms & 320ms & 400ms & 560ms & 640ms & 720ms & 880ms & 1000ms & Avg. 
\\ \hline\hline
PAML 
& 24.1 & 41.4 & 44.9 & 50.1 & 54.6
& 58.9 & 63.1 & 70.4 & 73.5 & 76.6 
& 55.8
& 26.4 & 35.5 & 47.0 & 59.4 & 71.1
& 81.4 & 87.3 & 97.0 & 109.8 & 123.5 
& 73.8\\
TimeHetNet 
& 27.4 & 45.6 & 49.1 & 53.5 & 58.6
& 66.4 & 69.3 & 74.5 & 77.1 & 81.9
& 60.3 
& 30.1 & 42.9 & 50.4 & 63.5 & 78.9 
& 86.7 & 93.2 & 104.6 & 123.8 & 139.6 
& 81.4 \\
MoPredNet 
& 20.4 & 32.1 & 35.4 & 40.4 & 43.1
& 50.5 & 53.0 & 54.9 & 57.2 & 60.7
& 44.8
& 18.0 & 29.8 & 37.5 & 48.2 & 64.7 
& 70.0 & 76.7 & 87.9 & 100.4 & 114.7 
& 64.8 \\
GraphHetNet 
& 21.2 & 33.7 & 36.2 & 38.6 & 39.7
& 48.8 & 50.2 & 51.1 & 53.4 & 55.9 
& 42.9 
& 21.2 & 33.6 & 41.9 & 51.0 & 67.0 
& 73.4 & 78.2 & 86.2 & 98.5 & 110.4 
& 66.1 \\
\hline
Ours 
& \textbf{15.5} & \textbf{20.2} & \textbf{22.6} & \textbf{26.0} & \textbf{30.3} 
& \textbf{37.6} & \textbf{39.6} & \textbf{40.5} & \textbf{40.9} & \textbf{42.9} 
& \textbf{31.6}
& \textbf{13.4} & \textbf{23.4} & \textbf{31.9} & \textbf{43.4} & \textbf{56.7} 
& \textbf{62.8} & \textbf{65.2} & \textbf{67.5} & \textbf{79.0} & \textbf{99.4} 
& \textbf{54.3}\\
\hline
\end{tabular}}
\resizebox{\textwidth}{!}
{
\begin{tabular}{c|cccccccccc|c|cccccccccc|c} \hline
\multirow{2}{*}{\textbf{Method}} 
& \multicolumn{11}{c|}{Basketball} 
& \multicolumn{11}{c}{Wash window} \\ 
\cline{2-23}
& 80ms & 160ms & 240ms & 320ms & 400ms & 560ms & 640ms & 720ms & 880ms & 1000ms & Avg. 
& 80ms & 160ms & 240ms & 320ms & 400ms & 560ms & 640ms & 720ms & 880ms & 1000ms & Avg. 
\\ \hline\hline
PAML 
& 28.4 & 39.5 & 52.7 & 64.2 & 81.6
& 92.3 & 107.7 & 115.9 & 123.4 & 131.7 
& 83.7
& 18.5 & 29.8 & 36.2 & 51.7 & 67.4 
& 78.7 & 84.9 & 93.4 & 97.2 & 101.7 
& 66.0\\
TimeHetNet 
& 29.5 & 43.4 & 57.0 & 68.1 & 84.8
& 96.4 & 110.1 & 117.4 & 126.7 & 139.0 
& 87.2
& 19.8 & 32.4 & 37.9 & 53.8 & 69.7 
& 82.1 & 86.4 & 92.1 & 94.5 & 99.1
& 66.8 \\
MoPredNet 
& 20.5 & 32.1 & 45.9 & 57.4 & 72.1
& 84.5 & 96.7 & 102.1 & 111.4 & 119.8 
& 74.3
& 12.5 & 20.1 & 26.9 & 41.2 & 50.8 
& 59.3 & 67.4 & 75.3 & 77.0 & 82.6 
& 51.3 \\
GraphHetNet 
& 21.8 & 34.5 & 46.2 & 58.8 & 73.7 
& 83.3 & 93.5 & 99.4 & 108.9 & 114.8 
& 73.5
& 14.5 & 21.9 & 27.6 & 43.2 & 54.2 
& 61.8 & 69.0 & 72.6 & 76.1 & 84.7
& 52.6\\
\hline
Ours 
& \textbf{14.7} & \textbf{28.4} & \textbf{40.2} & \textbf{52.1} & \textbf{63.2} & \textbf{76.4} & \textbf{83.5} & \textbf{88.2} & \textbf{98.7} & \textbf{104.1} & \textbf{65.0}
& \textbf{9.1} & \textbf{15.8} & \textbf{21.7} & \textbf{34.9} & \textbf{42.2} & \textbf{43.7} & \textbf{43.8} & \textbf{44.9} & \textbf{52.2} & \textbf{61.5} & \textbf{37.0}\\
\hline
\end{tabular}}
\vspace{-0.2cm}
\end{table*}

\section{Experiments}

To evaluate the effectiveness of our proposed method, we conduct experiments on the Human3.6M dataset \cite{ionescu2013human3} and the CMU Mocap dataset \footnote{Available at http://mocap.cs.cmu.edu/}.

\noindent\textbf{Human3.6M} \cite{ionescu2013human3} is a large-scale dataset popularly used in (few-shot) motion prediction. It contains 7 subjects performing 15 actions. On this dataset, following previous few-shot motion prediction works \cite{gui2018few,drumond2023few}, we evaluate our method on four actions (``walking'', ``eating'', ``smoking'', and ``discussion''), and use motion samples from the rest actions to form the base training set. Besides, during evaluation, we use the same testing motion sequences as existing methods \cite{gui2018few,drumond2023few}.

\noindent\textbf{CMU Mocap} is another large-scale human motion dataset. On this dataset, following previous few-shot motion prediction works \cite{zang2021few,zang2022few}, we use the motion sequences pre-processed by \cite{li2018convolutional}. Also following \cite{zang2021few,zang2022few}, we evaluate our method on four actions including ``running'', ``soccer'', ``basketball'', and ``wash window'', and use motion samples from the rest actions to form the base training set.

\noindent\textbf{Evaluation metric.} Following recent human motion prediction works \cite{xu2023auxiliary,guo2023back,gao2023decompose}, we evaluate our method using the Mean Per Joint Position Error (MPJPE) as the evaluation metric. On every frame (timestamp) of the future motion sequence, MPJPE measures the average L2 distance between the predicted joint coordinates and the ground-truth ones. As indicated in the recent motion prediction work \cite{xu2023auxiliary}, compared to Mean Angle Error (MAE) as a metric that is previously used, MPJPE can reflect a larger degree of human pose freedom, and can thus set up a clearer comparison between different methods.

\noindent\textbf{Implementation Details.} 
In FMP-OC, we use GPT-4 for ChatGPT. Moreover, for a fair comparison, following previous few-shot motion prediction works \cite{gui2018few,drumond2023few}, we set each observed motion sequence to be with 50 frames (2000ms), and set the number of supporting samples $N$ provided for each newly appearing human action to be 5. Besides, we set the size $P$ of the subset $S_{sub}$ to be 10. 
More details are in supplementary.

\subsection{Comparison with State-of-the-art Methods}

\begin{figure*}[h]
  \centering
  \vspace{1.2cm}
   \begin{subfigure}{\textwidth}
    \includegraphics[width=\linewidth]{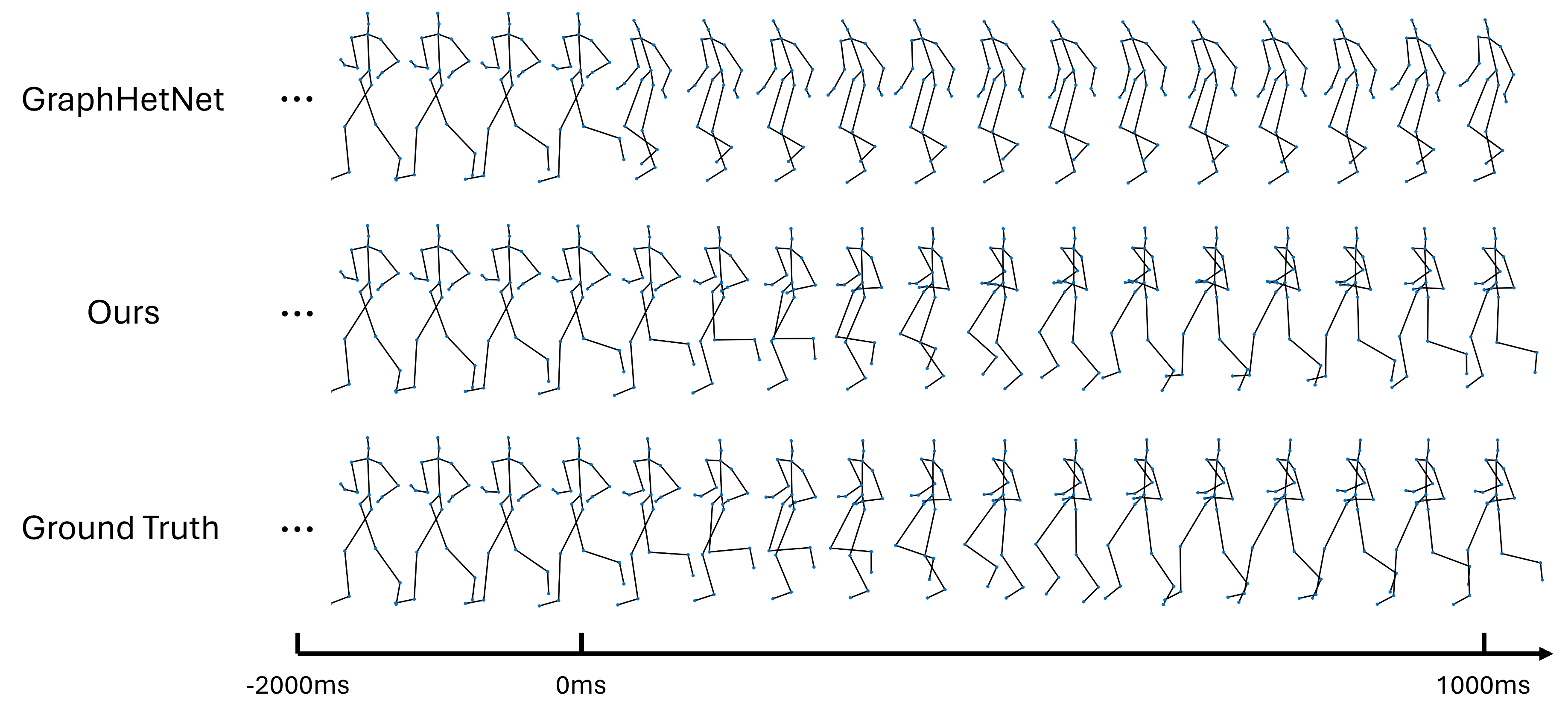}
    \caption{Results w.r.t the ``running'' action.}
    \vspace{0.3cm}
    \label{fig:visual_a2}
   \end{subfigure}
    \begin{subfigure}{\textwidth}
    \includegraphics[width=\linewidth]{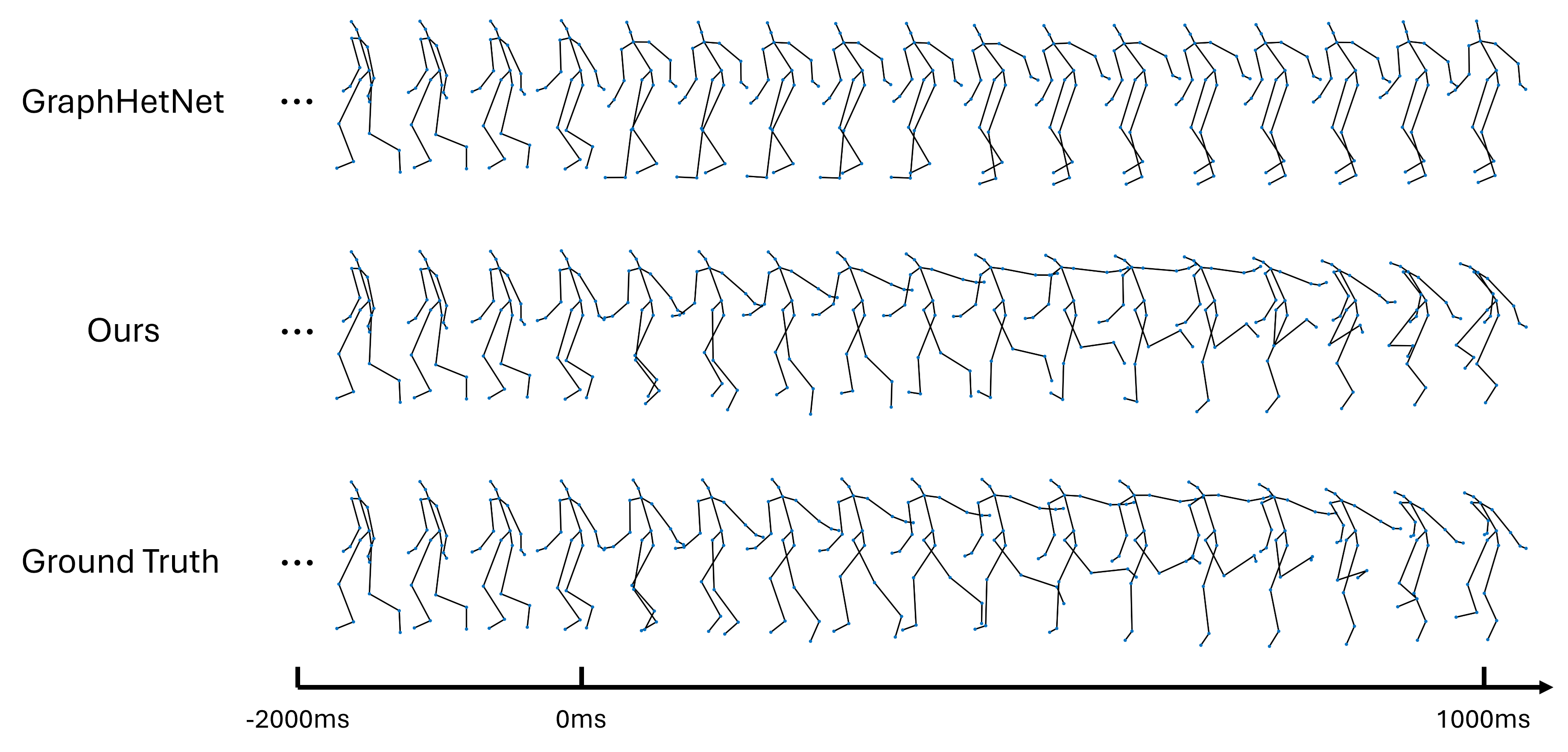}
    \caption{Results w.r.t the ``soccer'' action.}
    \label{fig:visual_b2}
   \end{subfigure}
   \caption{Qualitative results of our framework and the recent state-of-the-art few-shot human motion prediction method GraphHetNet \cite{drumond2023few} on the CMU Mocap dataset.} 
   \label{fig:visual_1}
\end{figure*}

\begin{figure*}[h]
  \centering
  \vspace{1.2cm}
   \begin{subfigure}{\textwidth}
    \includegraphics[width=\linewidth]{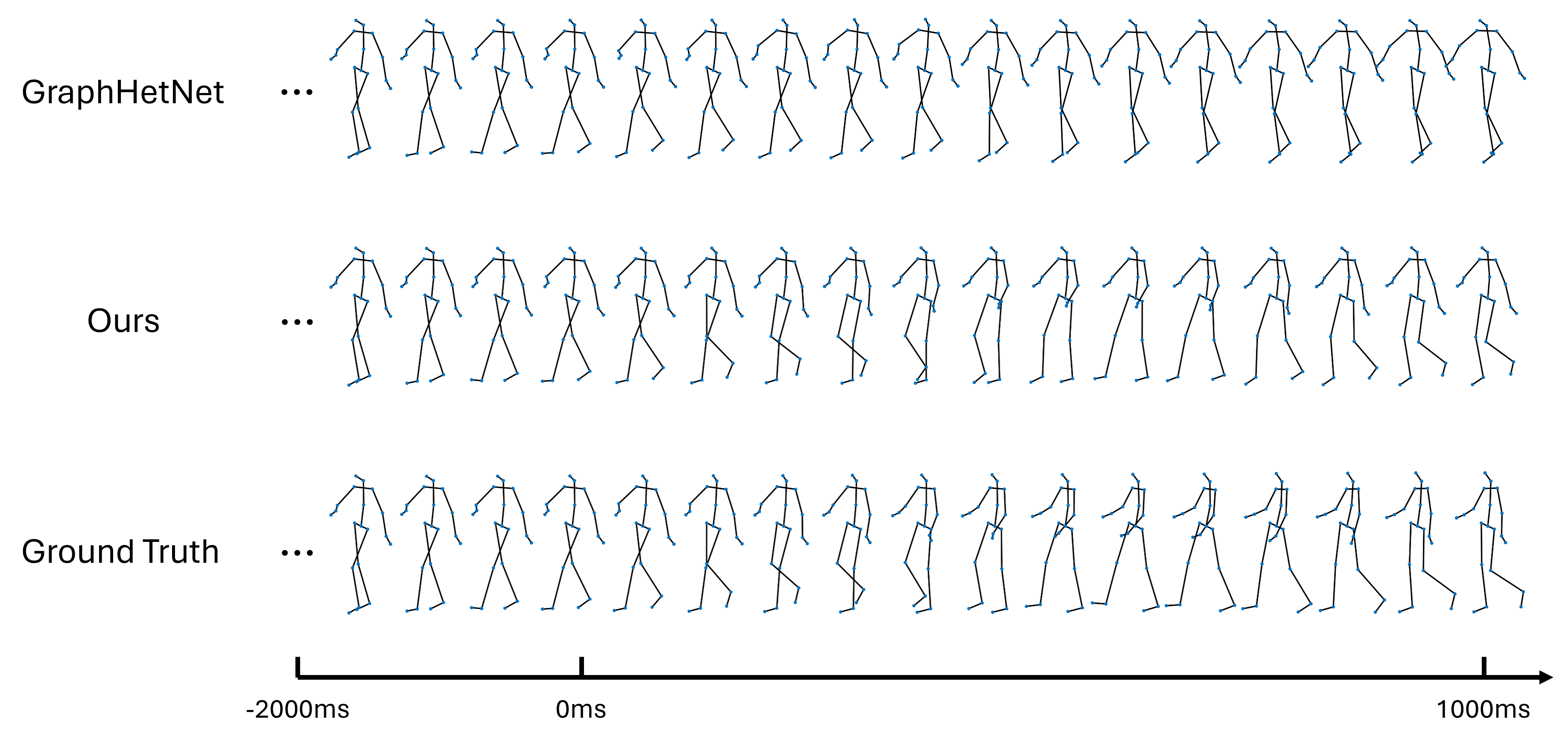}
    \caption{Results w.r.t the ``walking'' action.}
    \vspace{0.3cm}
    \label{fig:visual_a1}
   \end{subfigure}
    \begin{subfigure}{\textwidth}
    \includegraphics[width=\linewidth]{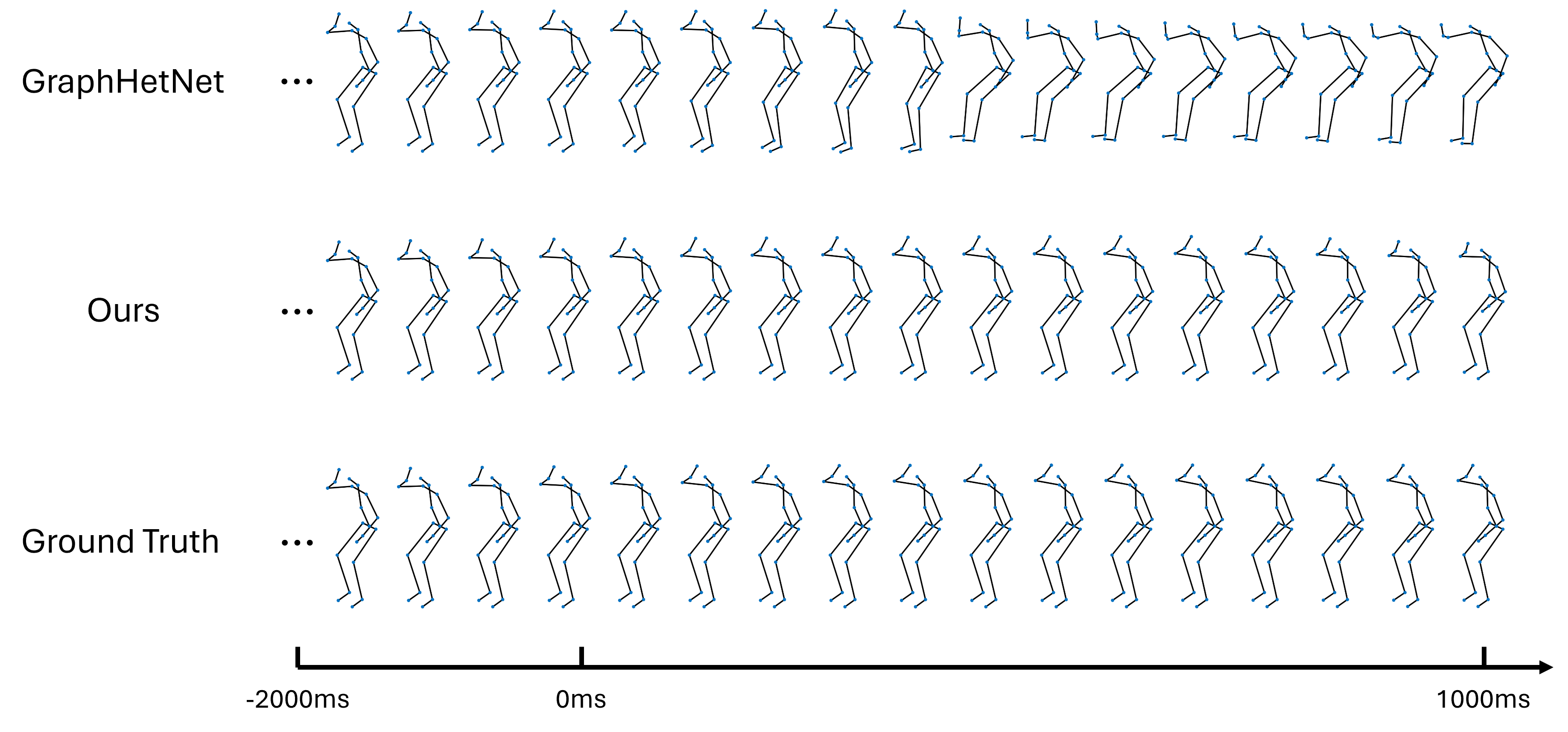}
    \caption{Results w.r.t the ``eating'' action.}
    \label{fig:visual_b1}
   \end{subfigure}
   \caption{Qualitative results of our framework and the recent state-of-the-art few-shot human motion prediction method GraphHetNet \cite{drumond2023few} on the Human3.6M dataset.} 
   \label{fig:visual_2}
\end{figure*}

In Tab.~\ref{Tab:human3.6m} and Tab.~\ref{Tab:cmu}, we report results on the Human3.6M dataset and the CMU Mocap dataset respectively.  On both datasets, we compare our approach with existing few-shot motion prediction methods including PAML \cite{gui2018few}, TimeHetNet \cite{brinkmeyer2022few}, MoPredNet \cite{zang2022few}, and GraphHetNet \cite{drumond2023few}. Moreover, on both datasets, for each action that is used for evaluation, we report (1) short-term motion prediction results (MPJPE at 80ms, 160ms, 240ms, 320ms, and 400ms), (2) long-term motion prediction results (MPJPE at 560ms, 640ms, 720ms, 880ms, and 1000ms), and (3) the average motion prediction result (MPJPE averaged over the above 5+5=10 timestamps). As shown, on all the actions used for evaluation across both datasets, in a totally training-free manner, our framework consistently surpasses all existing few-shot motion prediction methods by a large margin. This demonstrates the effectiveness of our framework. We also qualitatively compare our framework with the recent state-of-the-art method GraphHetNet \cite{drumond2023few}, and present results in Fig.~\ref{fig:visual_1} and Fig.~\ref{fig:visual_2}. As shown, 
compared to \cite{drumond2023few}, across both datasets, our framework consistently performs motion prediction more accurately. This further shows the efficacy of our framework. Note that we also convert these qualitative results into videos, and provide them in supplementary.

\subsection{Ablation Studies}

\textbf{Impact of the implicit knowledge extraction designs.} In this work, we involve two designs in our framework to facilitate extracting implicit knowledge from ChatGPT. (i.e., the sequence-description linkage design and the kinematic-chain-of-thought design). 
To evaluate their efficacy, we test three variants. In the first variant (\textbf{w/o both extraction designs}), we remove both designs from our framework. In the second variant (\textbf{w/o sequence-description linkage}), we still maintain the kinematic-chain-of-thought design but remove the other one. Moreover, in the third variant (\textbf{w/o kinematic-chain-of-thought}), we remove the kinematic-chain-of-thought design from our framework. As shown in Tab.~\ref{Tab:ablation_study_1}, our framework outperforms the first variant by a large margin. This shows the importance of extracting implicit knowledge from ChatGPT effectively, in guiding ChatGPT to be an accurate motion predictor. Furthermore, our framework also holds a better performance than both the second variant and the third variant. This further demonstrates the efficacy of the two implicit knowledge extraction designs we propose in our framework. \textbf{More
ablation studies such as the experiments w.r.t. the motion-in-context learning mechanism, the experiments w.r.t. the hyperparameter $P$, and the experiments w.r.t. other off-the-shelf large language models are in supplementary.}

\begin{table}[htbp]
\vspace{-0.25cm}
\centering
\caption{Evaluation on the implicit knowledge extraction designs.}
\vspace{-0.4cm}
\resizebox{\columnwidth}{!}
{
\small
\begin{tabular}{l|c|c|c|c}
\hline
Method & Walking & Eating & Smoking & Discussion\\
\hline
w/o both extraction designs & 45.3 & 56.1 & 42.8 & 81.9 \\
w/o sequence-description linkage & 42.1 & 51.6 & 38.9 & 78.0\\
w/o kinematic-chain-of-thought & 41.9 & 50.9 & 39.2 & 77.6\\
\hline
FMP-OC & 39.7 & 48.2 & 36.3 & 74.5 \\
\hline
\end{tabular}}
\label{Tab:ablation_study_1}
\vspace{-0.45cm}
\end{table}

\section{Conclusion}

In this paper, we have proposed a novel few-shot motion prediction framework FMP-OC. In this framework, instead of training a new motion prediction model, we for the first time, enable few-shot motion prediction as a non-language task to be performed directly via utilizing the off-the-shelf large language model ChatGPT, in a totally training-free manner.
Specifically, in FMP-OC, to acquaint ChatGPT as a language model with motion sequences, we first introduce several designs to effectively extract implicit knowledge from ChatGPT. Moreover, we also introduce FMP-OC with a motion-in-context learning mechanism, which further increases the familiarity of ChatGPT as a language model over the non-language motion prediction task.
Without requiring any training, our framework consistently outperforms the previous state-of-the-art methods by a large margin on the evaluated benchmarks. 
{
    \small
    \bibliographystyle{ieeenat_fullname}
    \bibliography{main}
}

\newpage

\end{document}